\documentclass[format=sigconf]{acmart}
\usepackage{booktabs} 
\usepackage{graphicx}
\usepackage{subfigure}
\usepackage{tikz}
\usepackage{amsmath}
\usepackage{amssymb}
\usepackage{multirow}
\usepackage{bm}
\usepackage{makecell}
\usepackage{pifont}
\usepackage{stackengine}
\usepackage{mathtools}

\newcommand{\orarrow}[1]{%
\overrightarrow{\vphantom{\mathbf{b}} {#1}}
}%

\newcommand{\olarrow}[1]{%
\overleftarrow{\vphantom{\mathbf{b}} {#1}}
}%

\newcommand{\dec}{
    \text{e}\widetilde{\text{m}}\text{b}
}

\copyrightyear{2019} 
\acmYear{2019} 
\setcopyright{acmlicensed}
\acmConference[KDD '19]{The 25th ACM SIGKDD Conference on Knowledge Discovery and Data Mining}{August 4--8, 2019}{Anchorage, AK, USA}
\acmBooktitle{The 25th ACM SIGKDD Conference on Knowledge Discovery and Data Mining (KDD '19), August 4--8, 2019, Anchorage, AK, USA}
\acmPrice{15.00}
\acmDOI{10.1145/3292500.3330900}
\acmISBN{978-1-4503-6201-6/19/08}

\title[QuesNet: A Unified Representation for Heterogeneous Test Questions]{QuesNet: A Unified Representation for\\Heterogeneous Test Questions}

\author { 
    Yu Yin$^1$, Qi Liu$^1$, Zhenya Huang$^1$, Enhong Chen$^{1,*}$, Wei Tong$^1$, Shijin Wang$^{2,3}$, Yu Su$^{2,3}$
}

\affiliation{
 \institution{$^1$Anhui Province Key Laboratory of Big Data Analysis and Application, School of Computer Science and Technology \& School of Data Science, University of Science and Technology of China,\\ \{yxonic,huangzhy\}@mail.ustc.edu.cn, \{qiliuql,cheneh\}@ustc.edu.cn, tongw@mail.neea.edu.cn}
}

\affiliation{
 \institution{$^2$iFLYTEK Research, iFLYTEK CO., LTD., $^3$State Key Laboratory of Cognitive Intelligence, \{sjwang3,yusu\}@iflytek.com}
}

\begin{document}

\fancyhead{}

\begin{abstract}
Understanding learning materials (e.g. test questions) is a crucial issue in online learning systems, which can promote many applications in education domain. Unfortunately, many supervised approaches suffer from the problem of scarce human labeled data, whereas abundant unlabeled resources are highly underutilized. To alleviate this problem, an effective solution is to use pre-trained representations for question understanding. However, existing pre-training methods in NLP area are infeasible to learn test question representations due to several domain-specific characteristics in education. First, questions usually comprise of heterogeneous data including content text, images and side information. Second, there exists both basic linguistic information as well as domain logic and knowledge. To this end, in this paper, we propose a novel pre-training method, namely QuesNet, for comprehensively learning question representations. Specifically, we first design a unified framework to aggregate question information with its heterogeneous inputs into a comprehensive vector. Then we propose a two-level hierarchical pre-training algorithm to learn better understanding of test questions in an unsupervised way. Here, a novel holed language model objective is developed to extract low-level linguistic features, and a domain-oriented objective is proposed to learn high-level logic and knowledge. Moreover, we show that QuesNet has good capability of being fine-tuned in many question-based tasks. We conduct extensive experiments on large-scale real-world question data, where the experimental results clearly demonstrate the effectiveness of QuesNet for question understanding as well as its superior applicability.

\end{abstract}

\ccsdesc[500]{Computing methodologies~Information extraction}
\ccsdesc[300]{Computing methodologies~Neural networks}
\ccsdesc[500]{Applied computing~Education}

\keywords{question representation, heterogeneous data, pre-training}

\maketitle

\section{Introduction}
In recent years, many online learning systems, such as Khan Academy and LeetCode, have gained more and more popularities among learners of all ages from K12, to college, and even adult due to its convenience and autonomy~\cite{moore2011distance,anderson2014engaging}. Holding large volume of question materials, these systems are capable of providing learners with many personalized learning experiences~\cite{ozyurt2013design}. 

In these platforms, it is necessary to well organize such abundant questions in advance~\cite{masud2012learning}. For example, we need to sort them by difficulty attributes or create curricula designs with their knowledge concepts. In practice, such managements are vitally necessary since they could help students save effort to locate required questions for targeted training and efficient learning~\cite{douglas2004effectiveness}. Therefore, it is essential to find an effective way for systems to understand test questions. In fact, since it is the fundamental issue promoting many question-based applications, such as difficulty estimation~\cite{huang2017question}, knowledge mapping~\cite{hermjakob2001parsing,zhang2003question} and score prediction~\cite{su2018exercise}, much attention has been attracted from both system creators and researchers.

In the literature, many efforts have been developed for understanding question content by taking advantage of natural language processing (NLP) techniques~\cite{hermjakob2001parsing,huang2017question}. In general, existing solutions usually design end-to-end frameworks, where the questions are represented as syntactic patterns or semantic encodings, and furthermore directly optimized in specific downstream tasks by supervised learning~\cite{huang2017question,tan2015lstm}. However, these task-specific approaches mostly require substantial amounts of manually labeled data (e.g., labeled difficulty), which restricts their performance in many learning systems that suffer from the sparsity problem of limited label annotations~\cite{huang2017question}. Comparatively, in this paper, we aim to explore an unsupervised way by taking full advantage of large-scale unlabeled question corpus available for question representation.

\begin{figure}
    \centering
    \includegraphics[width=0.3\paperwidth]{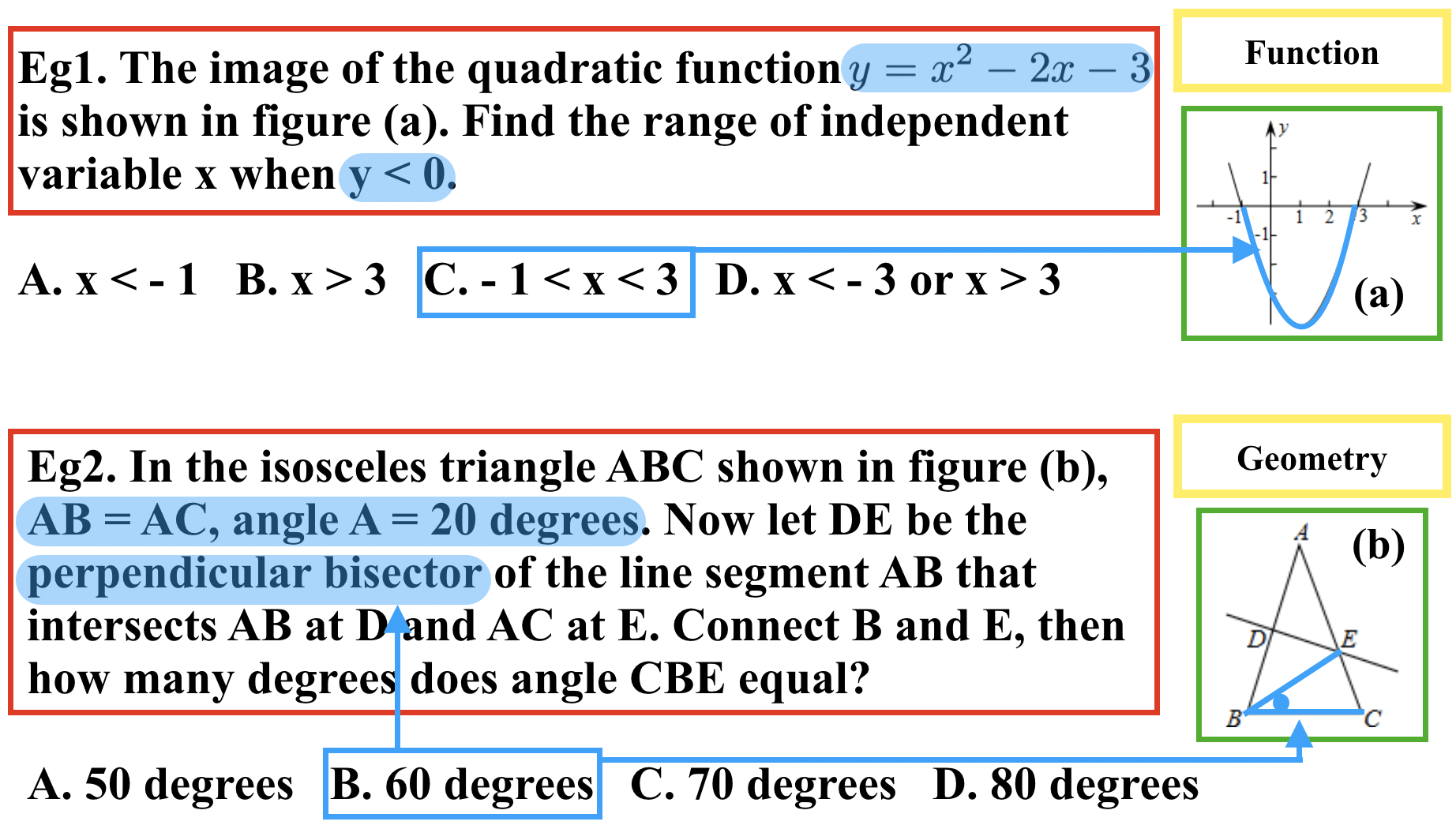}
    \caption{Two examples of heterogeneous questions.}\label{fig:case}
\end{figure}

Unfortunately, it is a highly challenging task. Although several pre-training methods have shown their superiority in NLP on tasks such as question answering~\cite{peters2018deep,devlin2018bert}, they just exploit the sentence context with homogeneous text. They are infeasible in understanding and representing question materials due to following domain-unique characteristics. First, test questions contain coherent heterogeneous data. For example, typical math questions in Figure~\ref{fig:case} comprise of multiple parts with different forms including text (red), image (green) and side information such as knowledge concept (yellow). All these kinds information are crucial for question understanding, which requires us to find an appropriate way to aggregate them for a comprehensive representation. Second, for a certain question, not only should we extract its basic linguistic context, but we also need to carefully consider the advanced logic information, which is a nontrival problem. As shown in Figure~\ref{fig:case}, in addition to linguistic context and relations from its content, a test question also contains high-level logic, taking the information of four options into consideration. The right answers are more related to the question meaning compared with the wrong ones, reflecting the unique mathematical logic and knowledge. E.g., to find the right answer (B) in question example 2, we need to focus more on the expression (``$AB=AC,\ldots$'') in text and the related $\angle CBE$ in the image. Third, in practice, the learned question representations should have great accessibility and be easy to apply to downstream tasks such as difficulty estimation. In actual educational tasks, question representations are often used as part of a complex model, which requires the method to have simple yet powerful structure and easy to mix-in into task-specific models.

To address the above challenges, in this paper, we provide a unified domain-specific method, namely QuesNet, for comprehensive learning test question representations. Generally, QuesNet is able to aggregate heterogeneous data of a certain question into an integrated form and gain a deeper understanding with the benefits of both low-level linguistic information and high-level domain logic knowledge. It can also be naturally applied to many downstream methods, which effectively enhances the performance on-the-fly. Specifically, we first design a unified model based on Bi-LSTM and self-attention structures to aggregate question information with its heterogeneous inputs into a vector representation. Then we propose a two-level hierarchical pre-training algorithm to learn better understandings of test questions. On the lower level, we develop a novel holed language model (HLM) objective to help QuesNet extract linguistic context relations from basic inputs (i.e., words, images, etc.). Comparatively, on higher level pre-training, we propose a domain-specific objective to learn advanced understanding for each question, which preserves its domain logic and knowledge. With objectives, QuesNet could learn the integrated representations of questions in an unsupervised way. Furthermore, we demonstrate how to apply QuesNet to various typical question-based tasks with fine-tuning in education domain including difficulty estimation, knowledge mapping and score prediction. We conduct extensive experiments on large-scale real world question data, with three domain-specific tasks. The experimental results clearly demonstrate that QuesNet has good capability of understanding test questions and also superior applicability of fine-tuning.


\section{Related Work}
We briefly summarize our related works as follows.

\subsection{Question Understanding}
Question understanding is a fundamental task in education, which have been studied for a long time~\cite{schwarz1996answering}. 
Generally, existing approaches could be roughly divided into two categories: rule-based representation and vector-based representation. For rule-based representation, scholars devote efforts to designing many fine-grained rules or grammars and learn to understand questions by parsing the question text into semantic trees or pre-defined features~\cite{graesser2006question,duan2008searching}. However, these initiative works heavily rely on expertise for designing effective rule patterns, which is obviously labor intensive. Comparatively, in vector-based representation methods, each question could be learned as a semantic vector in latent space automatically through many natural language processing (NLP) techniques~\cite{sundermeyer2012lstm,vaswani2017attention}. Recently, as an extension and combination of previous studies, deep learning techniques have become state-of-the-art models due to their superiority of learning complex semantics~\cite{huang2017question,zhang2018caden}. For example, Tan et al.~\cite{tan2015lstm} used Long Short-Term Memory (LSTM) model to capture the long-term dependency of question sentences. Huang et al.~\cite{huang2017question} utilized convolutional neural network for question content understanding, targeting at the difficulty estimation task. Although great success have been achieved, all these supervised methods suffer from the problem of scarce labeled data. That is, with labels only in specific task supervising both question understanding and task modeling pars, the understanding of the question is quite limited, while large volume of unlabeled question data bank is not leveraged. Moreover, none of the work have considered different question input forms, which causes an information loss for heterogeneous question understanding.

\begin{figure*}[ht]
    \centering
    \includegraphics[width=0.7\paperwidth]{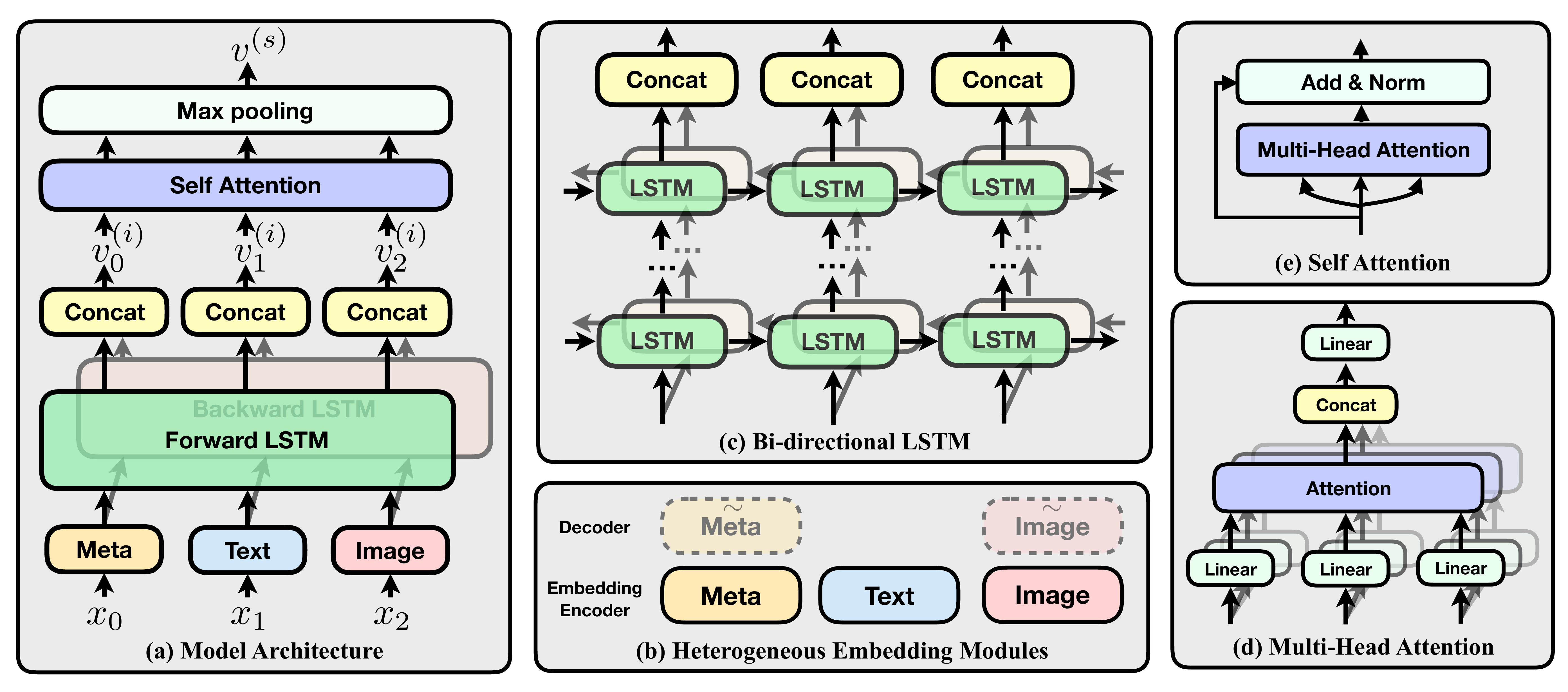}
    \caption{QuesNet model architecture. (a) shows the whole structure, which can be divided into three layers: 1. \emph{Embedding Layer}, with heterogeneous embedding modules in (b); 2. \emph{Content Layer}, with bi-directional LSTM detailed in (c); and 3. \emph{Sentence Layer}, which is a global self attention layer demonstrated in (e), the implementation shown in (d).}\label{fig:model}
\end{figure*}

\subsection{Text Pre-training}
Recent years have witnessed the development of pre-training methods, which is a good way to make the most of unlabeled corpus in NLP field~\cite{devlin2018bert}. These methods can be divided into two categories: feature-based methods, where text is represented by some sort of feature extractors as fixed vectors~\cite{pennington2014glove,peters2018deep}, and pre-training based methods, where parameters of model are pre-trained on corpus and then fine-tuned to specific tasks~\cite{howard2018universal,devlin2018bert}. Among them, the most successful model would be BERT~\cite{devlin2018bert}. It utilizes Transformer~\cite{vaswani2017attention} together with some language related pre-training goals, solving many NLP tasks with impressive performance. Although these pre-training solutions have been fully examined in a range of NLP tasks, yet they could hardly be directly applied to understanding test question mainly due to the following three reasons. First, test questions are heterogeneous, where much information exists in other formats, such as image and side features, would be ignored with these pre-training methods that only focus on text. Second, test questions contain much domain logic and knowledge to be understood and represented other than just linguistic features, which makes it hard for models to capture. Third, these approaches are difficult to be applied due to the need of model modification or hyper-parameter tuning, which is inconvenient under many education setup.

\subsection{Question-based Applications}
There are many question-based applications in education domain, which play important roles in traditional classroom setting or online learning system~\cite{anderson2014engaging}. The representative tasks include difficulty estimation~\cite{boopathiraj2013analysis,huang2017question}, knowledge mapping~\cite{desmarais2012item} and score prediction~\cite{piech2015deep}. Specifically, difficulty estimation requires us to evaluate how difficult a question is from its content without preparing a group of students to test on it. Knowledge mapping aims at automatically mapping a question to its corresponding knowledge points. Score prediction is a task of predicting how well a student performs on a specific question with their exercising history. All these applications benefit the system management and services, such as personalized recommendation~\cite{kuh2011piecing}.

Our work provides a unified representation for heterogeneous test questions compared with previous studies, and provides a solid backbone for applications in use of questions. We take the heterogeneous nature of test questions and the difficulty of understanding domain information into consideration and design a more powerful yet accessible pre-training algorithm. With heterogeneous question representation model and two-level pre-training, QuesNet captures much more information from test questions.


\section{QuesNet: Modeling and Pre-training}
In this section, we introduce QuesNet modeling and pre-training in detail. First we give a formal definition of the question representation problem. Then, we describe the QuesNet architecture for heterogeneous question representation. Afterwards we describe the pre-training process of QuesNet, i.e. the two-level pre-training algorithm. Finally, in Section~\ref{sec:fine_tuning}, we discuss how to apply QuesNet to downstream tasks and do fine-tuning.

\subsection{Problem Definition}
In this subsection, we formally introduce the question representation problem and clarify mathematical symbols in this paper.

In our setup, each test question $q$ is given as input in a heterogeneous form, which contains one or all kinds of content including text, images and side information (meta data such as knowledge). Formally, we could define it as a sequence $x=\{x_t\},t\in \{1, \ldots, T\}$, where $T$ is the length of the input question sequence, together with side information as a one-hot encoded vector $m\in\mathbf{R}^K$, where $K$ is the number of categories in side information. Each input item $x_t$ is either a word from a vocabulary (including formula piece), or an image in size $W\times H$.

For better usability, the final representation of each question $q$, i.e. the output, should contain both individual content representation as a sequence of vectors and the whole question representation as a single vector. We will see in Section~\ref{sec:fine_tuning} why all these representations are necessary. With the setup stated above, more formally, we define the question representation problem as follows:

\begin{definition}({\small \textbf{Question Representation Problem}}).
Given a question $q$ with heterogeneous input, as a sequence $\bm{x}=\{x_0,x_1,\ldots,x_T\}$, with side information as $x_0=m$, sequence length as $T$, each input content as $x_i$ (either a word or an $W\times H$ image), our goal is to represent $q$ as a sequence of content representation vectors $\bm{v}^{(i)}=\{v^{(i)}_0,\ldots,v^{(i)}_T\}$ and one sentence representation vector $v^{(s)}$, each of which is of dimension $N$. The representation should capture as much information as possible.
\end{definition}

In the following sections, we will address the main three challenges: (1) how QuesNet generates question representation; (2) how the representation is pre-trained; (3) how the representation is applied to downstream tasks.

\subsection{QuesNet Model Architecture}\label{sec:arch}
QuesNet model maps a heterogeneous question $q$ to a unified final representation $(\bm{v}^{(i)},v^{(s)})$. The architecture is shown in Figure~\ref{fig:model} (a), which can be seen as three layers: Embedding Layer, Content Layer and Sentence Layer. Specifically, given a question $q$, in Embedding Layer, its heterogeneous input embedding is performed. Then in Content Layer, Bi-LSTM is used to model different input content and generate each content representation $\bm{v}^{(i)}$. Finally, in Sentence Layer, we use self-attention to combine vectors in an effective way.

\subsubsection{Embedding Layer} We first introduce the building blocks of the Embedding Layer. The aim of this layer is to project heterogeneous input content to a unified space, which enables our model for different input forms. In order to do so, in first layer, we setup embedding modules to map each kind of inputs into fixed length vectors. Embedding module for words is a mapping $\text{emb}_w$ with parameters as $\theta_{we}$, which directly maps each word in the vocabulary to a vector of size $N_e$. Image embedding module $\text{emb}_i$, with parameters denoted as $\theta_{ie}$, as shown in the upper part of Figure~\ref{fig:model} (b), consists of three convolutional layers followed by activations. Features are also max-pooled into a vector of size $N_e$. Meta data embedding module $\text{emb}_m$, with parameters denoted as $\theta_{me}$, as shown in Figure~\ref{fig:model} (b), uses two layers of fully-connect neural network to embed input meta data as a vector of size $N_e$.

With these basic embedding modules, for each input item $x_t$ in $q$, we generate an embedded vector $e_t$ in the first layer, so that we can get an embedding sequence $\bm{e}=\{e_0,e_1,\ldots,e_T\}$ from input $\bm{x}=\{x_0,x_1,\ldots,x_T\}$. Formally:
\begin{equation*}
e_t=\left\{
\begin{array}{ll}
\text{emb}_w(x_t;\theta_{we}), & \text{if }x_t\text{ is word,} \\
\text{emb}_i(x_t;\theta_{ie}), & \text{if }x_t\text{ is image,} \\
\text{emb}_m(x_t;\theta_{me}), & \text{if }x_t\text{ is meta data.}
\end{array}
\right.
\end{equation*}

\subsubsection{Content Layer} In this layer, we aim at modeling relation and context for each input item. Existing methods like LSTM~\cite{hochreiter1997long} only cares about context on one side, while in Transformer~\cite{vaswani2017attention}, context and relation modeling relies on position embedding, which loses some locality. Therefore, with embedded vector sequence $\bm{e}$ described above as input, we incorporate a multi-layer bi-directional LSTM structure~\cite{huang2015bidirectional}, which is more capable of gaining context information. Here, we choose Bi-LSTM because it can make the most of contextual content information of question sentence from both forward and backward directions~\cite{huang2015bidirectional,ma2016end}. Specifically, given question embedding sequence $\bm{e}=\{e_0,e_1,\ldots,e_T\}$, we set the input of the first layer of LSTM as $\orarrow{\bm{h}}^{(0)}=\olarrow{\bm{h}}^{(0)}=\{e_0,e_1,\ldots,e_T\}$. At each position $t$, forward hidden states $(\orarrow{h}^{(l)}_t,\orarrow{c}^{(l)}_t)$ and backward hidden states $(\olarrow{h}^{(l)}_t,\olarrow{c}^{(l)}_t)$ at each layer $l$ are updated with input from previous layer $\olarrow{h}^{(l-1)}_{t}$ or $\olarrow{h}^{(l-1)}_{t}$ for each direction, and previous hidden states $(\orarrow{h}^{(l)}_{t-1},\orarrow{c}^{(l)}_{t-1})$ for forward direction or $(\olarrow{h}^{(l)}_{t+1},\olarrow{c}^{(l)}_{t+1})$ for backward direction in a recurrent formula as:
\begin{equation*}
\begin{array}{ll}
\orarrow{h}^{(l)}_{t},\orarrow{c}^{(l)}_{t}&=\text{LSTM}(\orarrow{h}^{(l-1)}_{t},\orarrow{h}^{(l)}_{t-1},\orarrow{c}^{(l)}_{t-1};\orarrow{\bm{\theta}}_{\text{LSTM}}),\\
\olarrow{h}^{(l)}_{t},\olarrow{c}^{(l)}_{t}&=\text{LSTM}(\olarrow{h}^{(l-1)}_{t},\olarrow{h}^{(l)}_{t+1},\olarrow{c}^{(l)}_{t+1};\olarrow{\bm{\theta}}_{\text{LSTM}}),
\end{array}
\end{equation*}
where recurrent formula follows Hochreiter et al.~\cite{hochreiter1997long}

More layers are needed for modeling deeper relations and context. With $L$ Bi-LSTM layers, deep linguistic information is able to be captured in hidden states. As hidden state at each direction only contains one-side context, it is beneficial to combine the hidden state of both directions into one vector at each step. Therefore, we obtain content representation at each time step $t$ as:
\begin{equation*}
v^{(i)}_{t}=\text{concatenate}(\orarrow{h}^{(L)}_{t},\olarrow{h}^{(L)}_{t}).
\end{equation*}

\begin{figure*}[ht]
    \centering
    \includegraphics[width=0.7\paperwidth]{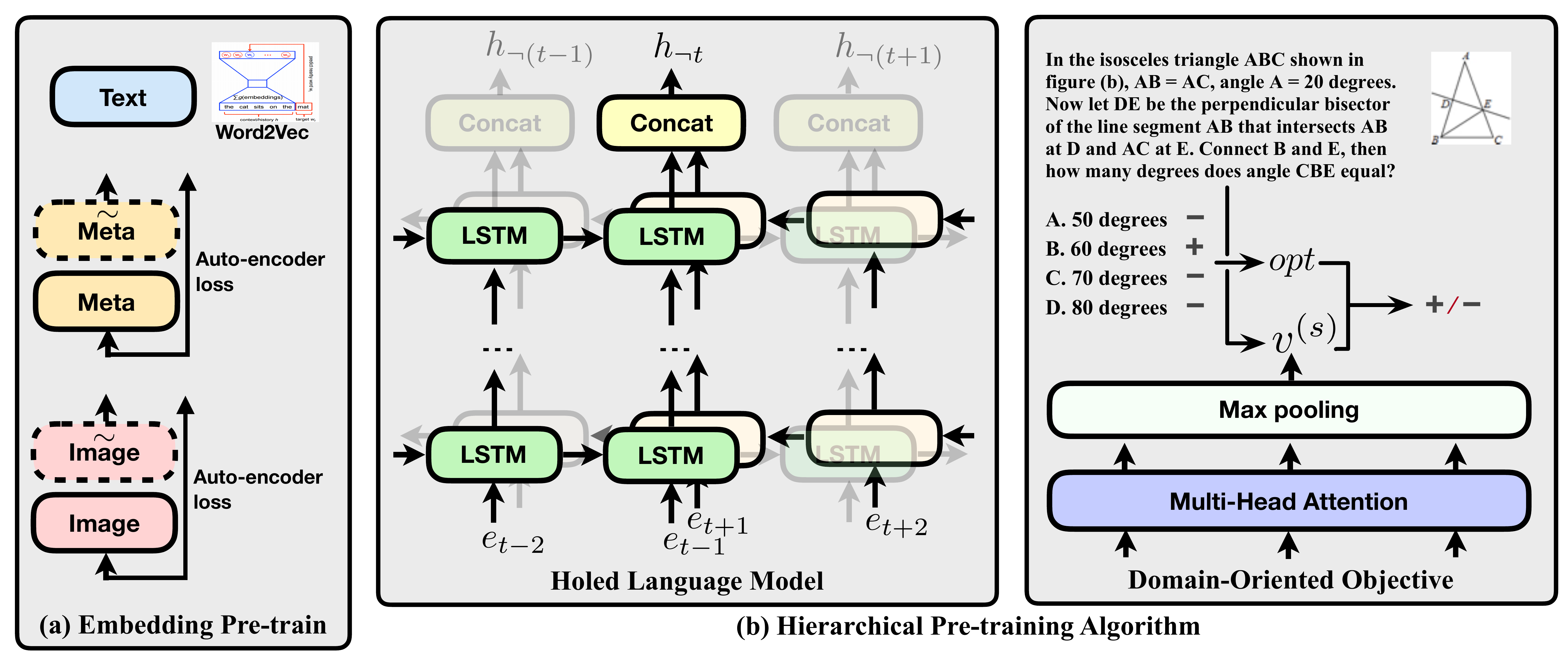}
    \caption{Pre-training of QuesNet. For pre-training, first, as shown in (a), \emph{Embedding Pre-training} is done in advance. Then in (b), two hierarchical objectives are defined: the low-level objective, Holed Language Model (HLM, middle), and the high-level domain-oriented objective (right).}\label{fig:algo}
\end{figure*}

\subsubsection{Sentence Layer} After we model lower-level linguistic features, we still have to aggregate this information in a way that focus on long-term and global complex relations, so that domain logic and knowledge can be captured. To this end, the third Sentence Layer consists of a self-attention module for aggregating item representation vector sequence $\bm{v}^{(i)}$ into a sentence representation $v^{(s)}$. Compared with LSTM that focuses on context, attention mechanism is more capable of modeling long-term logic and global information~\cite{vaswani2017attention}. Following Vaswani et al.~\cite{vaswani2017attention}, we use Multi-Head Attention module to perform global self attention. Given a set of queries of dimension $N$ (as matrix) $Q$, keys (as matrix) $K$, values (as matrix) $V$, Multi-Head Attention computes output matrix as:
\begin{equation*}
\begin{array}{ll}
\text{MultiHead}(Q,K,V)&=\text{concatenate}(\text{head}_1,\ldots,\text{head}_H)W^{O},\\
\text{head}_{j}&=\text{Attention}(QW_j^{Q},KW_j^{K},VW_J^{V}),\\
\text{Attention}(Q,K,V)&=\text{softmax}(QK^T/\sqrt{N})V,
\end{array}
\end{equation*}
where $H$ is the number of attention heads, $W^{O},W_j^{Q},W_j^{K},W_j^{V}$ are some projection matrices. Intuitively, Multi-Head Attention module performs several different attentions in parallel, which helps it to aggregate high level logic and knowledge for lower layers.

Within our setup, we use self-attention to aggregate content vectors $\bm{v}^{(i)}$ together with position embedding $pe^{(i)}$, by setting $Q$, $K$, $V$ in Multi-Head Attention all as $\bm{v}^{(i)}_+=\text{concatenate}(\bm{v}^{(i)}, pe^{(i)})$, and then attended values in all time step into one single vector with max-pooling. More formally:
\begin{equation*}
v^{(s)}=\max\left\{\text{MultiHead}(\text{LayerNorm}(\bm{v}^{(i)}_+,\bm{v}^{(i)}_+,\bm{v}^{(i)}_+) + \bm{v}_+^{(i)}))\right\},
\end{equation*}
where LayerNorm refers to layer normalization technique proposed by Ba et al.~\cite{ba2016layer}, and position embedding follows Vaswani et al.\cite{vaswani2017attention}

Till now we already generated a unified representation of a question. To summarize, with embedding layer, we embed heterogeneous content into a unified form. Then with multi-layer Bi-LSTM in content layer, we capture deep linguistic relation and context. Finally in sentence layer, we aggregate the information into a single vector with high level logic and knowledge.

\subsection{Pre-training}
However, we still need a way to learn all the linguistic features and domain logic on large unlabeled question corpus, which we will describe in this subsection. Specifically, we fully describe how to pre-train QuesNet to capture both linguistic features and domain logic and knowledge from question corpus. For this purpose, as shown in Figure~\ref{fig:algo}, we design a novel hierarchical pre-training algorithm. We first separately pre-train each embedding modules. Then in the main pre-training process, we propose two level hierarchical objectives. At low level pre-training, we proposed a novel holed language model as the objective for learning low-level linguistic features. At high-level learning, a domain-oriented objective is added for learning high-level domain logic and knowledge. The objectives of both levels are learned together within one pre-training process.

\subsubsection{Pre-training of Embedding} We first separately pre-train each embedding module to set up better initial weights for them. For word embedding, we incorporate \emph{Word2Vec}~\cite{mikolov2013efficient} on the whole corpus to get an initial word to vector mapping. For image and side information embedding, we first construct decoders for each embedding that decodes the vector given by embedding module. Then we train these embedding modules using auto-encoder losses~\cite{ngiam2011multimodal}. If we take image embedding module $emb_i$ with parameters $\theta_{ie}$ as an example, we first construct image decoder $\dec_i$ also with trainable parameters $\theta_{id}$. Then on all images in the corpus $\bm{I}$, we can construct auto-encoder loss as:
\begin{equation*}
\mathcal{L_I}=\sum_{x\in \bm{I}} l(\dec_i(\text{emb}_i(x)), x),
\end{equation*}
where $l(y,x)$ is a loss function that measures distance between y and x, such as mean-squared-error (MSE) loss. Then initial weights of image embedding module would be:
\begin{equation*}
\hat{\theta}_{ie} = \text{arg}\max_{\theta_{ie}}\mathcal{L_I}.
\end{equation*}

We can train initial values for side information embedding similarly. With all side information as $\mathcal{L_M}$, side information decoder $\dec(\cdot;\theta_{md})$ is implemented as a multi-layer fully-connected neural network, and the initial weights of it would be:
\begin{equation*}
\hat{\theta}_{me} = \text{arg}\max_{\theta_{me}}\mathcal{L_M}.
\end{equation*}

\subsubsection{Holed Language Model} The pre-training objective at low level aims at learning linguistic features from large corpus. Language model (LM) being the most used unsupervised linguistic feature learning objective, is limited by its one-directional nature. In this paper, we proposed a novel holed language model (HLM) that jointly combines context from both sides. Intuitively, the objective of HLM is to fill up every word with both left and right side context of it. It is different from the bi-directional LM implementation in ELMo~\cite{peters2018deep} where context from both sides are trained separately without any interaction. And it does not rely on random masking of tokens as BERT does, which is much more sample efficient.

In HLM, comparative to traditional language model, the probability of input content at each position $t$ is conditioned by its context at both sides, and our objective is to maximize conditioned probability at each position. Formally speaking, for an input sequence $\bm{x}$, the objective of HLM calculates:
\begin{equation*}
\mathcal{L}_{\text{HLM}}=-\sum_{t=0}^{T}\log P(x_t|x_{\lnot t}),
\end{equation*}
where we use $x_{\lnot t}$ to stand for all other inputs that are not $x_t$, and the goal is to minimize $\mathcal{L}_{\text{HLM}}$ (sum of negative log likelihood). As described in Section~\ref{sec:arch}, inputs on the left of position $t$ are modeled in the adjacent left hidden vector $\orarrow{h}_{t-1}$, on the right are modeled in the adjacent right hidden vector $\olarrow{h}_{t+1}$. Therefore, the conditional probability of each input item in HLM can be modeled using these two vectors combined:
\begin{equation*}
h_{\lnot t}=\text{concatenate}(\orarrow{h}_{t-1},\olarrow{h}_{t+1})
\end{equation*}
along with a succeeding output module, and a specific loss function compatible with negative log likelihood in the original.

Due to heterogeneous input of a test question, we have to model each input kind separately. For words, the output module would be a fully-connected layer $\text{out}_w$ with parameters $\theta_{ow}$, and the loss function would be cross entropy. The output module takes $h_{\lnot t}$ as input, and generates a vector of vocabulary size, which, after Softmax, models the probability of each word at position $t$. For images, the output module is a fully-connected layer $\text{out}_{i}(\cdot;\theta_{oi})$, followed by the image decoder $\dec(\cdot;\theta_{id})$ described before, and the loss function is mean-squared-error (MSE) loss. Next, the output module of side information is also a fully-connected layer $\text{out}_{m}(\cdot;\theta_{om})$, followed by the image decoder $\dec(\cdot;\theta_{md})$. The loss function of this kind of input is also cross entropy. Therefore, applying these output modules and loss functions above, the HLM loss at each position $t$ would become:
\begin{equation*}
l_t=\left\{
\begin{array}{ll}
\text{CrossEntropy}(\text{out}_w(h_{\lnot t}),x_t),& \text{if $x_t$ is a word;}\\
\text{MSE}(\dec_i(\text{out}_i(h_{\lnot t})),x_t),& \text{if $x_t$ is an image;}\\
\text{MSE}(\dec_m(\text{out}_m(h_{\lnot t})),x_t),& \text{if $x_t$ is side information.}
\end{array}\right.
\end{equation*}
Therefore, the objective of holed language model in the low level for question $q$ is 
\begin{equation*}
    \mathcal{L}_{\text{low}}=\sum_{t=0}^{T}l_t.
\end{equation*}

\subsubsection{Domain-Oriented Objective} The low level HLM loss only helps the model to learn linguistic features such as relation and context. However, domain logic and knowledge is still ignored. Take a look back at Figure~\ref{fig:case}, we can see that the relation between content and options contains much domain specific logic and knowledge. In order to also include such information in final representation, in this section, we designed a domain-oriented objective for high-level pre-training. We use the natural domain guidance of a test question: its answer and options, to help train QuesNet representation. For questions with one correct answer and other false options, we setup a pre-training task for QuesNet that, given an option, the model should output whether it is the correct answer. More specifically, we encode the option with a typical text encoder $\text{enc}(\cdot;\theta_{\text{opt}})$ and get the answer representation as $v_{\text{opt}}=\text{enc}(opt)$ where $opt$ represents the option. Then we model the probability of the option being the correct answer as:
\begin{equation*}
P(opt|q)=\text{sigmoid}\left(D(v^{(s)}_q,v_{\text{opt}};\theta_D)\right),
\end{equation*}
where $v^{(s)}_q$ is the sentence representation for $q$ generated by QuesNet, and $D(\cdot,\theta_D)$ is a fully-connected neural network with output of 1 dimension. Therefore, the domain-oriented objective in the high level for question $q$ is 
\begin{equation*}
    \mathcal{L}_{\text{high}}=-\log P(opt|q).
\end{equation*}

With both low and high level objective, the pre-training process can be conducted on a large question bank with heterogeneous questions. With embedding modules' weights initialized and pre-trained separately, we can now apply stochastic gradient descent algorithm to optimize our hierarchical pre-training objective:
\begin{equation*}
    \mathcal{L}=\mathcal{L}_{\text{low}}+\mathcal{L}_{\text{high}}.
\end{equation*}

After pre-training, QuesNet question representation should be able to capture both low-level linguistic features and high-level domain logic and knowledge, and transfer the understanding of questions to downstream tasks in the area of education.

\subsection{Fine-tuning}\label{sec:fine_tuning}
Downstream tasks in the area of education are often rather complicated. Taking knowledge mapping for example, as in the research by Yang et al.~\cite{yang2016hierarchical}, the authors use a fine-grained model on this multi-label problem, which requires representation for each input content. Another example is score prediction. In the paper of Su et al.~\cite{su2018exercise}, each exercise (test question) is represented as a single vector and then serves as the input of a sequence model.

As we can see, different tasks require different question representations. To apply QuesNet representation to a specific task, we just have to provide the required representation to replace the equivalent part of the downstream model, which minimizes the cost of model modification. Moreover, on each downstream task, only some fine-tuning of QuesNet is needed, which leads to faster training speed and better results.

In summary, QuesNet has the following advantages for question understanding. First, it provides a unified and universally applicable representation for heterogeneous questions. Second, it is able to learn not only low-level linguistic features such as relation and context, but also high-level domain logic and knowledge. Third, it is easy to apply to downstream tasks and do fine-tuning. In next section, we will conduct extensive experiments to further demonstrate these advantages.

\section{Experiments}
In this section, we conduct extensive experiments with QuesNet representation on three typical tasks in the area of education to demonstrate the effectiveness of our representation method. 

\subsection{Experimental Setup}\label{sec:setup}

\subsubsection{Dataset} 
\begin{figure}
    \centering
    \includegraphics[width=0.4\paperwidth]{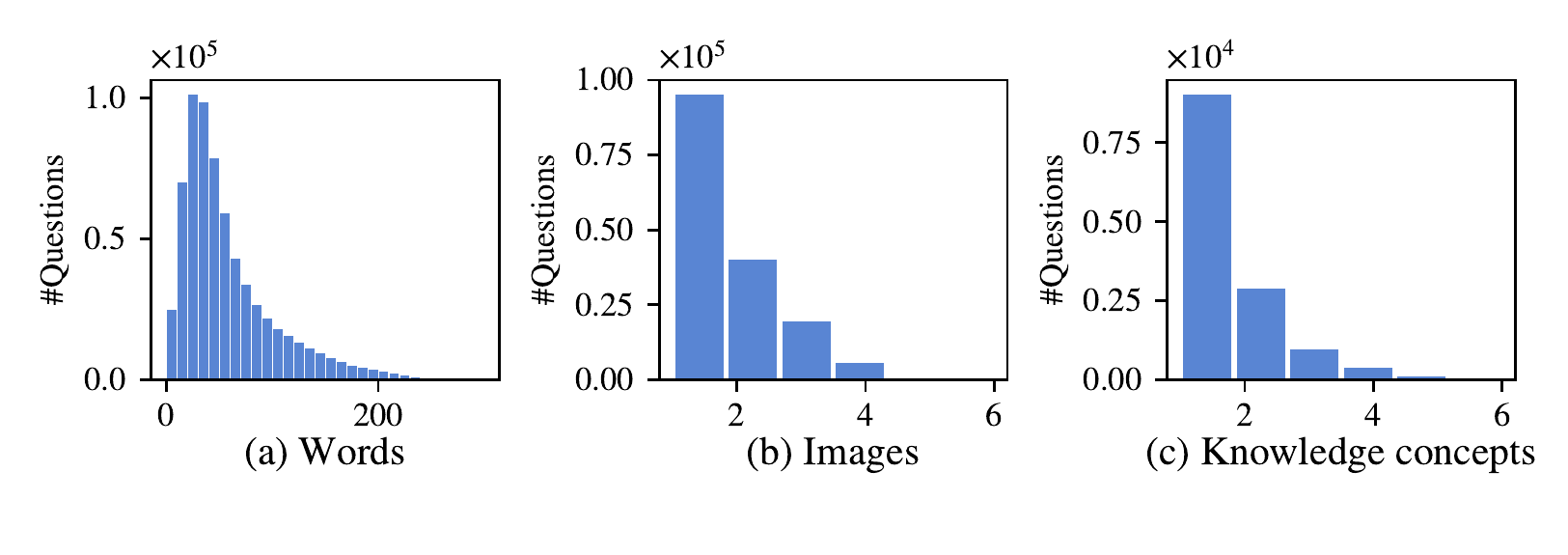}
    \caption{Distribution of question inputs and labels.}\label{fig:stat}
\end{figure}

The dataset we used, along with the large question corpus, are supplied by iFLYTEK Co., Ltd., from their online education system called Zhixue\footnote{http://www.zhixue.com}. All the data are collected from high school math tests and exams. Some important statistics are shown in Table~\ref{tab:dataset} and Figure~\ref{fig:stat}. The dataset is clearly heterogeneous, as shown in the table. About 25\% of questions contain image content, and about 72\% of questions contain side information. To clarify, side information used in knowledge mapping task is some other question meta data (its grade, the amount of which is shown in Table~\ref{tab:dataset}). In all other tasks, knowledge concepts are used as side information. Ignoring the heterogeneous information would definitely cause a downgrade in question understanding. We also observed that questions contain about 60 words in average, but the information contained is much more, as there are plenty of formulas represented as \LaTeX\ in question text.

\subsubsection{Evaluation Tasks} We pick three typical tasks related with test questions in the area of education, namely: knowledge mapping, difficulty estimation, and student performance prediction. The unlabeled question corpus contains around 0.6 million questions. All of the questions in corpus are later used for pre-training each comparison models. For traditional text representation models, image and side information inputs are omitted.

The main objective for knowledge mapping task is to map a given question to its corresponding knowledge~\cite{piech2015deep}. This is a multi-label task, where about 13,000 questions are labeled (only 1.98\% of the whole unlabeled question dataset). To show how a representation method alleviate this scarce label problem and how it performs on this task, we choose a state-of-the-art knowledge mapping model, and replace its question representation part with each representation model we want to compare. After fine-tuning, we use some of the mostly used evaluation metrics for multi-label problem including accuracy (ACC), precision, recall, and F-1 score. Details of these metrics can be found in Piech et al.~\cite{piech2015deep}, Yang et al.~\cite{yang2016hierarchical}.

The second task, namely difficulty estimation, is a high-level regression task to estimation the difficulty of a question. The scarce problem is even worse, in that merely 0.37\% of the questions have been labeled. Meanwhile, the task needs more domain logic and knowledge as a guidance to get better performance, as estimation of the difficulty of an exercise requires a deeper understanding of the question. The dataset on this task consists of only 2400 questions. The evaluation metrics, following Huang et al.~\cite{huang2017question}, includes Mean-Absolute-Error (MAE), Root-Mean-Squared-Error (RMSE), Degree of Agreement (DOA), and Pearson Correlation Coefficient (PCC).

The score prediction task, on the other hand, is a much more complicated domain task, where the main goal is to sequentially predict how well a student performs on each test question they exercises on~\cite{su2018exercise}. While student record is of large scale, the amount of questions for this task is still quite limited, only about 2.22\%. For better modeling student exercising sequence, some of the state-of-the-art model incorporate question content combining a question representation into the modeling. We replace this module with our comparison methods, and evaluate the performance using MAE, RMSE (mentioned earlier), accuracy (ACC) and Area Under the Curve (AUC), as used in many studies~\cite{zhang2017dynamic,su2018exercise}.

\begin{table}
    \caption{Statistics of datasets.} \label{tab:dataset} 
    \begin{tabular}{l|cccc}
      \toprule
       & All & KM & DE & SP \\
      \midrule
      \#Questions & 675,264 & 13,372 & 2,465 & 15,045 \\
      \#Questions with image & 165,859 & 3,318 & 299 & 2,952 \\
      \#Questions with meta & 488,352 & 8,030 & 1,896 & 5,948 \\
      \#Questions with option & 242,960 & 4840 & 1,389 & 4,364\\
      Avg. Words per question & 59.10 & 58.43 & 60.25 & 51.94 \\
      \#Students & - & - & - & 50,945 \\
      \#Records & - & - & - & 3,358,111 \\
      Label sparsity & - & 1.98\% &0.37\%&2.22\%\\
      \bottomrule
    \end{tabular}
  \end{table}

\subsubsection{QuesNet Setup}\footnote{The code is available at https://github.com/yxonic/QuesNet.} The embedding modules all output vectors of size 128. The image embedding module and related decoder are implemented as 4 layer convolutional neural networks and transposed convolutional neural networks, respectively. The size of feature maps at each layer is 16, 32, 32, 64. The side information embedding module and related decoder are setup as two-layer fully-connected neural networks. The size of the hidden layer is set to 256. For the main part of QuesNet, we use $L=4$ layer of bi-LSTM and 1 layer of self-attention. Sizes of hidden states in these modules are set to 256. To prevent overfitting, we also introduce dropout layers~\cite{srivastava2014dropout} between each layer, with dropout probability as 0.2.

Before any pre-training, the layers are first initialized with Xavier initialization strategy~\cite{glorot2010understanding}. Then at pre-training process, parameters are updated by Adam optimization algorithm~\cite{kingma2014adam}. We pre-train our model on question corpus long enough so that the pre-training loss converges. For optimizers in each task, we follow the setups described in the corresponding paper.

\subsubsection{Comparison Methods} We compare QuesNet with several representation methods. All these methods are able to generate question representation, and then be applied to the three evaluation tasks mentioned above. Specifically, these methods are:
\begin{itemize}
    \item \textbf{Original} refers to original supervised learning methods on each of the evaluation task. We choose state-of-the-art methods for difficulty estimation~\cite{huang2017question}, knowledge mapping~\cite{yang2016hierarchical} and score prediction~\cite{su2018exercise}, without any forms of pre-training.
    \item \textbf{ELMo} is a LSTM based feature extraction method with bidirectional language model as pre-training strategy~\cite{peters2018deep}. It is only capable of text representation, so we omit other types of input when using this method.
    \item \textbf{BERT} is a state-of-the-art pre-training method featuring Transformer structure and masked language model~\cite{devlin2018bert}. It is only capable of text representation, so we also omit other types of input.
    \item \textbf{H-BERT} is a modified version of BERT, which allows it to process heterogeneous input. We use the same input embedding modules as QuesNet, and set the embedded vectors as the input of text-oriented BERT.
\end{itemize}

\begin{table}
\caption{Comparison methods.} \label{tab:methods} 
\begin{tabular}{l|ccc|cc}
    \toprule
    Method & Text & Image & Meta & Low level & High level \\
    \midrule
    Original & \ding{52} & & & -&-\\
    ELMo & \ding{52} & & & -&-\\
    BERT & \ding{52} & & & -&-\\
    H-BERT & \ding{52} & \ding{52} & \ding{52} & - & - \\
    QN-T & \ding{52} & & & \ding{52}&\ding{52}\\
    QN-I & & \ding{52} & & \ding{52}&\ding{52}\\
    QN-M & & & \ding{52} & \ding{52}&\ding{52}\\
    QN-TI & \ding{52} & \ding{52} & & \ding{52}&\ding{52}\\
    QN-TM & \ding{52} & & \ding{52} & \ding{52}&\ding{52}\\
    QN-IM &  & \ding{52} & \ding{52} & \ding{52}&\ding{52}\\
    QN (no pre) & \ding{52} & \ding{52} & \ding{52} & & \\
    QN-L & \ding{52} & \ding{52} & \ding{52} & \ding{52} &\\
    QN-H & \ding{52} & \ding{52} & \ding{52} & & \ding{52}\\
    QuesNet & \ding{52} & \ding{52} & \ding{52} & \ding{52} & \ding{52}\\
    \bottomrule
\end{tabular}
\end{table}

\begin{table*} 

    \caption{Performance of comparison methods on different tasks.} \label{tab:performance} 
    \centering
    \begin{tabular}{l|c|c|c|c|c|c|c|c|c|c|c|c} 
        \toprule
        \multirow{2}{*}{Methods} &\multicolumn{4}{c}{Knowledge mapping}\vline & \multicolumn{4}{c}{Difficulty estimation}\vline & \multicolumn{4}{c}{Student performance prediction}\\
        \cline{2-13} & ACC & Precision & Recall & F-1 & MAE & RMSE & DOA & PCC & MAE & RMSE & ACC & AUC \\
        \midrule
        Original & 0.5744 & 0.4147 & 0.7872 & 0.5432 & 0.2200 & 0.2665 & 0.6064 & 0.3050 & 0.4245 & 0.4589 & 0.7459 & 0.5400 \\
        ELMo & 0.6942 & 0.7960 & 0.7685 & 0.7820 & 0.2250 & 0.2655 & 0.5561 & 0.4299 & 0.3569 & \textbf{0.4361} & 0.7866 & 0.5773\\
        BERT & 0.6224 & 0.7326 & 0.6711 & 0.7005 & 0.2265 & 0.2975 & 0.6258 & 0.3600 & 0.4009 & 0.4630 & 0.7390 & 0.5279 \\
        H-BERT & 0.6261 & 0.7608 & 0.6911 & 0.7243 & 0.2097 & 0.2698 & \textbf{0.6597} & 0.3713 & 0.3925 & 0.4528 & 0.7784 & 0.5838 \\
        QuesNet & \textbf{0.7749} & \textbf{0.8659} & \textbf{0.8075} & \textbf{0.8357} & \textbf{0.2029} & \textbf{0.2530} & 0.6137 & \textbf{0.4499} & \textbf{0.3445} & 0.4403 & \textbf{0.7999} & \textbf{0.6354} \\
        \bottomrule
    \end{tabular}
\end{table*}

\begin{table*} 

    \caption{Ablation experiments.} \label{tab:ablation} 
    \centering
    \begin{tabular}{l|c|c|c|c|c|c|c|c|c|c|c|c} 
        \toprule
        \multirow{2}{*}{Methods} &\multicolumn{4}{c}{Knowledge mapping}\vline & \multicolumn{4}{c}{Difficulty estimation}\vline & \multicolumn{4}{c}{Student performance prediction}\\
        \cline{2-13} & ACC & Precision & Recall & F-1 & MAE & RMSE & DOA & PCC & MAE & RMSE & ACC & AUC \\
        \midrule
        QN-T & 0.7050 & 0.8264 & 0.7436 & 0.7829 & 0.2166 & 0.2733 & 0.6123 & 0.3040 & 0.4488 & 0.4713 & 0.7454 & 0.6052 \\
        QN-I & 0.1136 & 0.2232 & 0.3195 & 0.2628 & 0.2265 & 0.2713 & 0.5961 & 0.2178 & 0.4711 & 0.4899 & 0.7400 & 0.5921 \\
        QN-M & 0.0355 & 0.1396 & 0.2853 & 0.1875 & 0.2251 & 0.2737 & 0.5549 & 0.2205 & 0.4719 & 0.4908 & 0.7410 & 0.5502 \\
        QN-TI & 0.7207 & 0.8307 & 0.7595 & 0.7935 & 0.2110 & 0.2647 & 0.6029 & 0.3333 & 0.4279 & 0.4678 & 0.7523 & 0.6221 \\
        QN-TM & 0.7196 & 0.8428 & 0.7523 & 0.7950 & 0.2114 & 0.2664 & 0.6151 & 0.3315 & 0.4353 & 0.4803 & 0.7456 & 0.6156 \\
        QN-IM & 0.1428 & 0.2323 & 0.2818 & 0.2547 & 0.2277 & 0.2707 & 0.5766 & 0.2279 & 0.4710 & 0.4906 & 0.7411 & 0.5513 \\
        QN (no pre) & 0.5659 & 0.6816 & 0.7091 & 0.6951 & 0.2225 & 0.2657 & 0.5750 & 0.3087 & 0.4349 & 0.4759 & 0.7488 & 0.5891 \\ 
        QN-L & 0.7185 & 0.8352 & 0.7457 & 0.7879 & 0.2193 & 0.2630 & 0.5721 & 0.3359 & 0.3843 & 0.4561 & 0.7747 & 0.6237 \\
        QN-H & 0.6807 & 0.8052 & 0.7271 & 0.7642 & 0.2161 & 0.2665 & \textbf{0.6291} & 0.3328 & 0.3946 & 0.4475 & 0.7814 & 0.6058 \\
        QuesNet & \textbf{0.7749} & \textbf{0.8659} & \textbf{0.8075} & \textbf{0.8357} & \textbf{0.2029} & \textbf{0.2530} & 0.6137 & \textbf{0.4499} & \textbf{0.3445} & \textbf{0.4403} & \textbf{0.7999} & \textbf{0.6354} \\
        \bottomrule
    \end{tabular}
\end{table*}

All comparison methods are listed in Table~\ref{tab:methods}. For a fair comparison, all these methods are adjusted to contain approximately same amount of layers and parameters, and all of them are tuned to have the best performance. All models are implemented by PyTorch, and trained on a cluster of Linux servers with Tesla K20m GPUs.

\subsection{Experimental Results}\label{sec:performance}
The comparison results on each of three tasks with four different models including QuesNet are shown in Table~\ref{tab:performance}. We can easily see that pre-trained methods is able to boost the performance on each task, among which QuesNet has the best performance on almost all metrics, no matter what the size of each task is. This proves that QuesNet gains a better understanding of questions and is transfered more efficiently from large unlabeled corpus to small labeled datasets. However, there are more to be explained in this table. First, models that support heterogeneous input have superior results over similar structures without heterogeneous input, which proves that it is crucial to handle heterogeneous inputs when understanding questions. Second, as all methods are adjusted to contain similar amount of parameters, QuesNet turns out to be the most efficient one. Third, the result of the Transformer based methods is slightly lower than other LSTM based methods. This is probably because of the low sample efficiency of masked language model pre-training strategy used in BERT, while bi-directional language model in ELMo performs better, and our novel holed language model used in QuesNet outperforms both of them.

\subsection{Ablation Experiments}\label{sec:ablation}
In this section, we conduct some ablation experiments to further show how each part of our method affect final results. In Table~\ref{tab:ablation}, there are eight variations of QuesNet, each of which takes out one or more opponents from the full QuesNet. Specifically: \emph{QN-T}, \emph{QN-I}, \emph{QN-M} refer to QuesNet with only text, image or side information is used in pre-training process, respectively. \emph{QN-TI}, \emph{QN-TM}, \emph{QN-IM} refer to combinations of different kinds of input, i.e. text and image, text and side information, and image and side information, respectively. And finally, \emph{QN-L} refers to QuesNet that only includes low-lever pre-training objective (holed language model), and \emph{QN-H} refers to QuesNet that only includes high-level domain objective.

The result in Table~\ref{tab:ablation} indeed shows many interesting conclusions. First, the more information a model incorporates, the better the performance, which agrees with the intuition. Second, if we focus on comparison between QN-H and QN-L, we will notice that they gain different effects on different tasks. On tasks more focusing on lower level features like knowledge mapping, QN-L outperforms QN-H slightly, while on other more domain-oriented high-level tasks (difficulty estimation and student performance prediction), QN-H performs a little better. This clearly demonstrates different aspects the two objectives focus on, and with full QuesNet outperforms both QN-L and QN-H. we know that QuesNet is able to take account of both aspects of these objectives, and build an understanding at both low linguistic level and high logic level. Third, we notice that QN-T has better performance than QN-I, which is better than QN-M, indicating that text carries most information in a question, then images, then side information. But as omitting either of the input kind will cause a performance loss, all of the information in heterogeneous inputs is essential to a good question understanding.

\subsection{Discussion}
From the above experiments, it is clear that QuesNet can effectively gain an understanding of test questions. First, it can well aggregate information from heterogeneous inputs. The model is able to generate unified representation for each question with different forms, and leverage information in all kinds of input. Second, with low-level objective capturing linguistic features, and high-level objective learning domain logic and knowledge, QuesNet is also able to gain both low-level and high-level understanding of test questions. Impressive performance of QuesNet can be seen among all three typical educational applications, which highlights the usability and superiority of QuesNet in the area of education.

There are still some directions for future studies. First, we may work on some domain specific model architectures to model logic among questions in a more fine-grained way. Second, the understanding of QuesNet on test questions is not comprehensible, and in the future, we would like to work on comprehension and explanation, to generate a more convincing representation. Second, the general idea of our method is a applicable in more heterogeneous scenarios, and we would like to further explore the possibilities of our work on other heterogeneous data and tasks.


\section{Conclusion}

In this paper, we presented a unified representation for heterogeneous test questions, namely QuesNet. Specifically, we first designed a heterogeneous modeling architecture to represent heterogeneous input as a unified form. Then we proposed a novel hierarchical pre-training framework, with holed language model (HLM) for pre-training low-level linguistic features, and a domain-oriented objective for learning high-level domain logic and knowledge. With extensive experiments on three typical downstream tasks in education from the low-level knowledge mapping task, to the domain-related difficulty estimation task, then to complex high-level student performance prediction task, we proved that QuesNet is more capable of question understanding and transferring, capturing both low-level linguistic features and high-level domain logic and knowledge. We hope this work builds a solid basis for question related tasks in the area of education, and help boost more applications in this field.

\section*{Acknowledgements}

This research was partially supported by grants from the National Key Research and Development Program of China (No. 2016YFB1000904) and the National Natural Science Foundation of China (Grants No. 61727809, U1605251, 61672483). Qi Liu gratefully acknowledges the support of the Young Elite Scientist Sponsorship Program of CAST and the Youth Innovation Promotion Association of CAS (No. 2014299).

\bibliographystyle{ACM-Reference-Format}
\bibliography{kdd}
\end{document}